\documentclass[letterpapers, 10 pt, conference]{ieeeconf}
\IEEEoverridecommandlockouts                                                          

\usepackage{cite}
\usepackage{amsmath,amssymb,amsfonts}
\usepackage{gensymb}
\usepackage{graphicx}
\usepackage{textcomp}
\usepackage{xcolor}
\usepackage{verbatim}
\usepackage{afterpage}
\usepackage{cite}
\usepackage{overpic}
\usepackage{psfrag}
\usepackage{latexsym}
\usepackage{bm}
\usepackage{cases}
\usepackage{array}
\usepackage{fancyhdr}
\usepackage{setspace}
\usepackage{subfigure}
\usepackage{url}
\usepackage{multirow}
\usepackage{epstopdf}
\usepackage{epsfig}
\usepackage{fancybox}
\usepackage{textcomp}
\usepackage{physics}
\usepackage{romannum}

\usepackage[ruled,linesnumbered]{algorithm2e}

\DeclareMathOperator*{\argminA}{arg\,min} 
\newcommand{\x}[0]{\mathbf{x}}

\newcommand{\rot}[0]{\mathbf{R}}
\newcommand{\tran}[0]{\mathbf{t}}
\newcommand{\SE}[0]{\text{SE}}
\newcommand{\SO}[0]{\text{SO}}
\newcommand{\R}[0]{\mathbb{R}}
\newcommand{\ranging}[0]{\mathbf{r}}

\begin{document}

\title{Distributed Ranging SLAM for Multiple Robots with Ultra-WideBand and Odometry Measurements}

\author{Ran Liu, Zhongyuan Deng, Zhiqiang Cao, Muhammad Shalihan, Billy Pik Lik Lau, Kaixiang Chen, \\Kaushik Bhowmik, Chau Yuen, and U-Xuan Tan
\thanks{This work is supported by National Key R\&D Program of China (2019YFB1310805) and National Science Foundation of China (12175187).}
\thanks{R. Liu, M. Shalihan, B. P. L. Lau, K. Bhowmik, C. Yuen, and U-X. Tan are with the Singapore University of Technology and Design, Singapore 487372. 
\tt\small \{ran\_liu, billy\_lau, kaushik\_bhowmik, yuenchau, uxuan\_tan\}@sutd.edu.sg}
\thanks{Z. Deng, Z. Cao, and K. Chen are with the Southwest University of Science and Technology, Mianyang, Sichuan, China 621010.}
}

\maketitle

\begin{abstract}
To accomplish task efficiently in a multiple robots system, a problem that has to be addressed is Simultaneous Localization and Mapping (SLAM).
LiDAR (Light Detection and Ranging) has been used for many SLAM solutions due to its superb accuracy, but its performance degrades in featureless environments, like tunnels or long corridors.
Centralized SLAM solves the problem with a cloud server, 
which requires a huge amount of computational resources and lacks robustness against central node failure.
To address these issues, 
we present a distributed SLAM solution to estimate the trajectory of a group of robots using Ultra-WideBand (UWB) ranging and odometry measurements. 
The proposed approach distributes the processing among the robot team 
and significantly mitigates the computation concern emerged from the centralized SLAM.
Our solution determines the relative pose (also known as loop closure) between two robots by minimizing the UWB ranging measurements taken at different positions when the robots are in close proximity. 
UWB provides a good distance measure in line-of-sight conditions, 
but retrieving a precise pose estimation remains a challenge, 
due to ranging noise and unpredictable path traveled by the robot.
To deal with the suspicious loop closures, we use Pairwise Consistency Maximization (PCM) to examine the quality of loop closures and perform outlier rejections. 
The filtered loop closures are then fused with odometry in a distributed pose graph optimization (DPGO) module to recover the full trajectory of the robot team.
Extensive experiments are conducted to validate the effectiveness of the proposed approach.
\end{abstract}

\section{Introduction}
\label{sec:introduction}
Multiple robots system collectively accomplishes a task by the collaboration of a group of robots, which is shown to be efficient and robust when compared to single robot solution \cite{Liu2020Cooperative}. Localization is crucial for the robot team to coordinate with each other. 
The literature shows a number of mature techniques and implementations for localization in known infrastructures. However, a prior knowledge of the environment is not always available in some scenarios. For example, emergency response often requires a group of robots to explore unknown site within a minimal time. 

Therefore, it is crucial for a group of robots to realize Simultaneous Localization and Mapping (SLAM) in unknown environments. 
Most multiple robots SLAM systems are based on centralized or distributed architecture \cite{resource_allocaltion_swarm} \cite{dpgo}. 
Centralized solution processes all collected measurements in a server \cite{resource_allocaltion_swarm}.
The computational cost increases exponentially with the increase of the number of robots.
Distributed system outperforms centralized solution in scalability and computational efficiency, 
as the computation is allocated among a group of robots \cite{dpgo}.
We are interested in distributed SLAM solution for multiple robots during the exploration of unknown environments.

\begin{figure}
\centering
\includegraphics[width=0.35\textwidth]{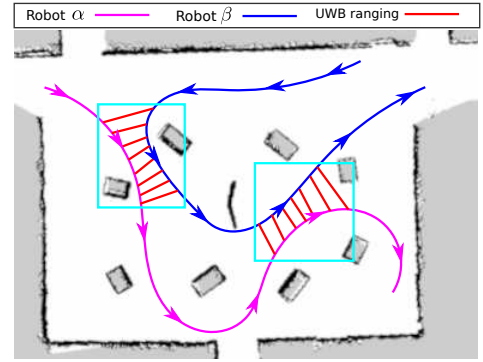}
\caption{Overview of the proposed distributed ranging SLAM with two robots. 
Each robot carries a UWB node for ranging of the nearby robots. 
We perform distributed pose estimation with short-term UWB ranging and odometry measurements (see the two cyan rectangles) when the two robots are in close proximity. The estimated poses are passed for outlier rejection and then optimized through a distributed pose graph optimization module. 
}
\vspace{-0.5cm}
\label{fig:motivation}
\end{figure}

SLAM has been extensively researched by the robotics community with LiDAR and visual-based sensors \cite{Kimera_Multi}, 
which exhibit good localization accuracy in general, but it is challenging to deal with featureless scenes (for example tunnels) and the algorithms often require a lot of computational resources. 
Due to its high accuracy and detection without line-of-sight, Ultra-WideBand (UWB) is widely used for localization \cite{Yassin2017Recent} \cite{Liu_relative_positioning_icra}.
Although a precise distance is provided by UWB, the lack of bearing information makes it difficult to be applied for the navigation of autonomous robots. 
The conventional Angle-of-Arrival (AoA)-based solution requires a cumbersome setup for phase shift estimation \cite{aoa_estimation}. 
We focus on the use of purely UWB ranging and odometry for pose estimation among a robot team.
The proposed solution is supposed to work in non-line-of-sight conditions, 
where LiDAR and visual sensors often fail to perform localization due to the block of the environment.

This paper presents a distributed SLAM for trajectory estimation of a group of robots based on UWB ranging and odometry measurements.
Each robot is equipped with a UWB node for ranging, as shown in Figure \ref{fig:motivation}. 
We estimate the pose (position and orientation) between the robots using short-term UWB ranging and odometry observations. 
In particular, we achieve the distributed pose estimation (i.e., loop closure) through the minimization of ranging measurements taken at different positions when two robots are close. 
The UWB is susceptible to non-line-of-sight (NLOS) propagation \cite{shalihan_case}, 
which results in erroneous loop closures, 
in particular when the path traveled by the robot is unpredictable \cite{nguyen2022relative}. 
To remove erroneous loop closures, 
we utilize the distributed Pairwise Consistency Maximization (PCM) algorithm to verify the quality of loop closure pairs and perform oulier rejections.
The inlier loop closures are finally incorporated with the odometry through Distributed Pose Graph Optimization (DPGO) to estimate the trajectory of the robots.
We evaluated the performance of the proposed approach with three robots in an indoor environment. 
The contributions of this paper are summarized as follows:
\begin{enumerate}
\item We propose an approach for distributed pose estimation between two robots using short-term UWB ranging and odometry measurements. In particular, we combine coarse search and optimization strategy to avoid the convergence to a local minimum.
\item We utilize distributed outlier rejection to remove the erroneous estimations and apply distributed pose graph optimization for trajectory estimation.
To the best of our knowledge, this is the first work that uses UWB ranging for distributed SLAM with multiple robots.
\item We perform extensive experiments to evaluate the performance of the proposed approach. The results show 
that we achieved a positioning accuracy of 0.45m in translation and 4.01$\degree$ in rotation by the fusing of UWB and odometry measurements in an indoor environment with a size of 10m$\times$12m.
\end{enumerate}

We organize the remaining of this paper as follows: Section \ref{sec:related_work} introduces the related work. 
Section \ref{sec:distributed_slam} describes the proposed distributed ranging SLAM. 
Section \ref{sec:experimental_results} shows the experimental setups and the experimental results. 
Finally, Section \ref{sec:conclusions} concludes this paper and discusses the future work.

\section{Related Work}
\label{sec:related_work}
The robotics research community is showing a growing interest in multiple robots system, 
due to the efficiency by collaboration of a robot team.
The system is extremely useful when multiple robots with different functionalities are used for exploration of unknown environments, 
like underground exploration in DARPA subterranean challenge\footnote[1]{https://www.subtchallenge.com/}. 
SLAM is considered fundamental to achieve this task due to its capability of localization and mapping in unknown environments.
Tremendous progress has been made concerning multiple robots SLAM with different sensors and implementations.

Visual cameras and LiDARs are popular for localization and SLAM in robotics, due to their high accuracy \cite{Kimera_Multi} \cite{LAMP_multipe_lidar_slam} \cite{LiDAR_visual_slam}. 
Kimera-multi \cite{Kimera_Multi} is presented for robust distributed trajectory estimation and 3D metric-semantic meshing in multiple robots applications.
Ebadi \textit{et al.} \cite{LAMP_multipe_lidar_slam} implemented a LiDAR-based SLAM and tested it for the exploration of underground tunnels with multiple robots. 
CamVox \cite{LiDAR_visual_slam} is a low-cost and robust SLAM solution that combines Livox LiDAR and visual features for robust sensor fusion.

However, the performance of LiDAR and visual SLAM decreases in perceptually-degraded environments, like tunnels and long corridors.
Due to detection without line-of-sight and high accuracy, UWB has been used for localization in many industrial applications. 
A sensor fusion scheme is presented in \cite{xielihua_uwb_visual_lidar} to combine visual, IMU, and UWB for accurate and low-drift localization.
Liu \textit{et al.} \cite{mapping_uwb_lidar} combined short-range LiDAR and UWB for mapping of mobile robots in indoor environments. 
Cao \textit{et al.} \cite{yanjun_cao_uwb} proposed a solution to localize a moving robot with a single UWB anchor based on the fusion of IMU with an Extended Kalman Filter.
Funabiki \textit{et al.} \cite{range_assisted_mapping} presented an approach to combine UWB beacon and LiDAR-based geometric loop closures for SLAM in large and perceptually challenging environments.

The implementation of multiple robots SLAM can be briefly classified into centralized and distributed. 
The centralized SLAM requires a server for data collection and sensor fusion. 
Cao \textit{et al.} \cite{Cao_iros2021} used four UWB nodes with a square configuration to determine the pose between two robots. 
But equipping multiple UWB sensors is costly and is not suitable for small robot applications.
Liu \textit{et al.} \cite{Liu2020Cooperative} proposed an approach to combine IMU and UWB for localization of multiple users. 
Martel \cite{relative_pose_linear} \textit{et al.} presented an approach for relative pose estimation with six ranging measurements based on linear algorithm.
Nguyen \textit{et al.} \cite{xielihua_uwb_collaborative} presented an approach to fuse UWB, IMU, and visual measurements to achieve multi-robot collaborative localization.
A centralized system is not scalable with large robot teams, 
due to the increase of computation and communication requirements.

In distributed SLAM, the team works collaboratively to achieve the goal of SLAM and the computation is distributed to peer robots. 
The Gauss-Seidel approach is presented in \cite{gauss_seidel_SLAM} for distributed trajectory estimation, which is shown to scale well to large robot teams. 
Tian \textit{et al.} \cite{dpgo} presented a distributed SLAM solution based on sparse semidefinite relaxation, which is guaranteed to be globally optimal under moderate measurement noise.
This approach is improved by \cite{shaojie_bdpgo} to balance the computational and communication cost.
Lajoie \textit{et al.} \cite{doorslam} presented DOOR-SLAM, which is a fully distributed SLAM system based on robust place recognition.
\begin{figure}
\centering
\includegraphics[width=0.45\textwidth]{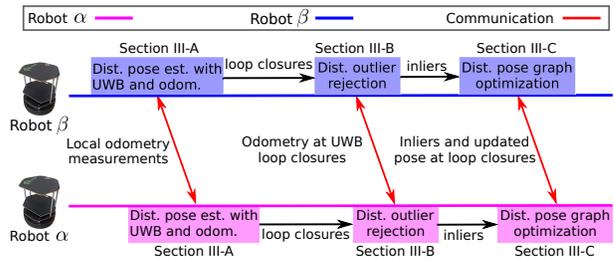}
\caption{Overview of the modules and their communication requirements for our distributed ranging SLAM with two robots.}
\label{fig:overview}
\vspace{-0.7cm}
\end{figure}
\section{Distributed Ranging SLAM for Multiple Robots}
\label{sec:distributed_slam}
In this section, we formulate the problem of distributed SLAM with UWB ranging and odometry for a group of robots. 
Then, we show the details of our proposed distributed SLAM, which mainly consists of the following three modules (see Figure \ref{fig:overview}). 
Firstly, each robot performs distributed pose estimation to determine the pose of its peer robots using short-term UWB ranging and odometry measurements.
Secondly, each robot runs a distributed outlier rejection module to filter out the suspicious estimations.
Thirdly, a distributed SLAM framework is used to fuse the inliers and the odometry 
to produce the trajectory in a common reference frame.

The robot team consists of $N$ robots and we denote the pose of robot $\alpha$ ($1 \le \alpha \le N$) at time $t$ as ${{\x}_\alpha^{t}} = [\rot_\alpha^{t}, \tran_\alpha^{t}] \in \SE(2)$, where $\rot_\alpha^{t} \in \SO(2)$ represents the rotation (i.e., $\theta_\alpha^{t}$) and $\tran_\alpha^{t} \in \R^2$ represents the 2D position (i.e., $x_\alpha^{t}$ and $y_\alpha^{t}$).
${{\x}_\alpha^{k,i}}$ denotes the relative pose from time $k$ to time $i$ measured by odometry of robot $\alpha$. 
We use $\ranging_\alpha^{t}=\{r_{\alpha, \beta}^{t}\}$ ($\beta \in N_{\alpha}^{t})$ to denote the UWB ranging measurements received by robot $\alpha$ at time $t$, where $N_{\alpha}^{t}$ denotes the set of neighboring robots of robot $\alpha$ and $r_{\alpha, \beta}^t$ represents the UWB ranging between robot $\alpha$ and robot $\beta$ at time $t$. 
An example of the problem can be illustrated in Figure \ref{fig:overview}, where two robots are performing distributed SLAM in unknown environment. 
Each robot carries a UWB node for ranging to its peer robots in close proximity. 
The robot also offers odometry to represent its relative movement. 
The goal is to achieve the trajectory estimation for multiple mobile robots through the UWB ranging and odometry measurements without a prior knowledge about the infrastructure. 
Our approach works in a distributed fashion, where each robot will perform its pose estimation by considering the measurements (i.e., odometry and UWB) received from all neighboring robots.  

\subsection{Module 1: Distributed Pose Estimation with Short-term UWB and Odometry Measurements}
\label{pose_est_ranging}
The estimation of the relative pose $\overline{\x}_{\alpha, \beta}^{t}$ (i.e., pose of robot $\beta$ with reference to robot $\alpha$ at time $t$) can be achieved by finding the best pose configuration $\overline{\x}_{\alpha,\beta}^{t}$ through minimizing the residual error of UWB ranging taken at different positions (denoted by the odometry) within a sliding time window $\tau$: 
\begin{equation}
\small
 \begin{split}
\argminA_{\x_{\alpha, \beta}^t} \sum_{i=t-\tau}^{i=t} \underbrace{\mathbf{e}(\x_{\alpha, \beta}^{t}, \x_{\alpha}^{t}, \x_{\beta}^{t}, \x_{\alpha}^{i}, \x_{\beta}^{i}, r_{\alpha,\beta}^{i})}_{\text{Residual error of UWB ranging}} \,\,\,\,\,\,\,\,\,\,\,\,\,\,\,\,\,\,\,\,\,\,\,\,\,\,\,\,\,\,\,\,\,\,&\\ 
=\argminA_{\x_{\alpha, \beta}^t} \sum_{i=t-\tau}^{i=t} (\underbrace{r_{\alpha,\beta}^{i}}_{\text{Ranging}} {-} || { \underbrace{{(\x_{\alpha}^{t})}^{{\text -1}}\cdot\x_{\alpha}^{i}}_{\text{Odom. of $\alpha$}}, \x_{\alpha, \beta}^t \cdot\underbrace{{(\x_{\beta}^{t})}^{{\text -1}}\cdot\x_{\beta}^{i}}_{\text{Odom. of $\beta$}}}|| )^2,
\label{eq:optimization}
 \end{split}
\end{equation}
where $\mathbf{e}(\cdot)$ denotes the residual function that computes the residual error of a UWB ranging given a pose configuration $\x_{\alpha, \beta}^t$. 
$||\cdot||$ computes the Euclidean distance between two positions. $\tau$ denotes the size of the sliding window for pose estimation.
The optimization of Equation \ref{eq:optimization} can be solved efficiently by state-of-the-art optimization tools, such as g2o \cite{g2o}, gtsam \cite{gtsam}, and Ceres \cite{ceres-solver}, if a good initial guess of $\x$ is provided. 
Due to the non-convexity of Equation \ref{eq:optimization},
the optimization converge to a local minimum without a proper initial guess and there is no guarantee to find the best solution. 

\begin{algorithm}
\small
\label{searching_algorithm}
\SetKw{KwWith}{with}
\SetKw{KwEach}{each}
\SetKw{KwBreak}{break}
  \SetKwData{Left}{left}\SetKwData{This}{this}\SetKwData{Up}{up}
  \SetKwFunction{Union}{Union}\SetKwFunction{FindCompress}{FindCompress}
  \SetKwInOut{Input}{input}\SetKwInOut{Output}{output}
\KwData{Odometry from robot $\alpha$: $\x_{\alpha}^i$, odometry from robot $\beta$: $\x_{\beta}^i$, UWB ranging between $\alpha$ and $\beta$: $\ranging_{\alpha, \beta}^i$, where $t-\tau \le i \le t$}
\KwResult{Coarse pose estimation $\widetilde{\x}_{\alpha,\beta}^t$}
\caption{Find coarse relative pose $\widetilde{\x}_{\alpha,\beta}^t$ between robot $\alpha$ and robot $\beta$ with short-term UWB and odom.}
\tcp{Record the minimum residual}
$minimumResidual \leftarrow \infty$ \\
  \For{$i_\phi \leftarrow -w_\phi$ \KwTo $w_\phi$}{
    \For{$i_\theta\leftarrow -w_\theta$ \KwTo $w_\theta$}{    
    $\x_{\alpha,\beta}' = <r_{\alpha, \beta}^t\cos(\delta i_\phi), r_{\alpha, \beta}^t\sin(\delta i_\phi), \delta i_\theta> $\\
    \tcp{Initialize the residual to be 0}
    $r \leftarrow 0 $ \\
    \For{$i \leftarrow t-\tau$ \KwTo $t$}{    
    \tcp{Residual with Equation \ref{eq:optimization}}
    $r = r+\mathbf{e}(\x_{\alpha, \beta}', \x_{\alpha}^{t}, \x_{\beta}^{t}, \x_{\alpha}^{i}, \x_{\beta}^{i},r_{\alpha,\beta}^{i})$\\
    \tcp{Bound residual error for fast computation}
    \If{$r \ge minimumResidual$}{\KwBreak} 
    }   
   \If{$r<minimumResidual$}
   {$minimumResidual \leftarrow r $\\
   $\widetilde{\x}_{\alpha, \beta}^t = <x_{\alpha,\beta}', y_{\alpha,\beta}', \theta_{\alpha,\beta}'>  $\\
   }
    }
  }
\Return $\widetilde{\x}_{\alpha, \beta}^t$
\end{algorithm}

\begin{figure}
  \centering
  \subfigure[Odom. trajectories for Case \Romannum{1}]{
\label{fig:trajectory_case1}
        \includegraphics[width=0.22\textwidth]{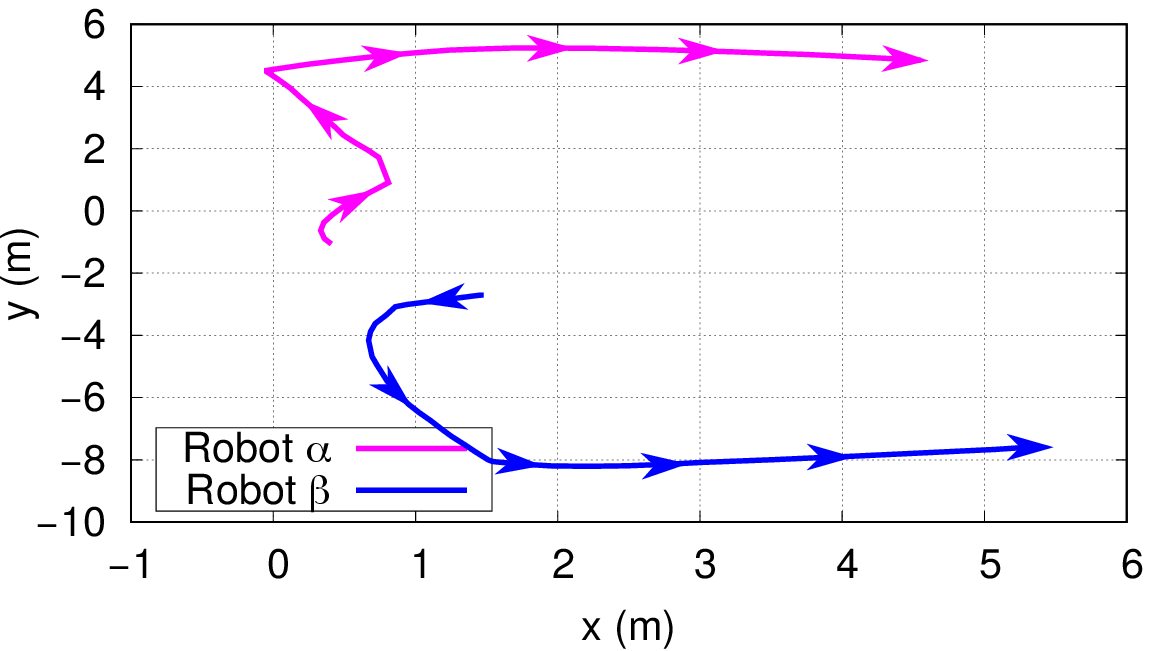}
        }
\hspace{-0.05in}\subfigure[Odom. trajectories for Case \Romannum{2} ]{
\label{fig:trajectory_case2}
        \includegraphics[width=0.22\textwidth]{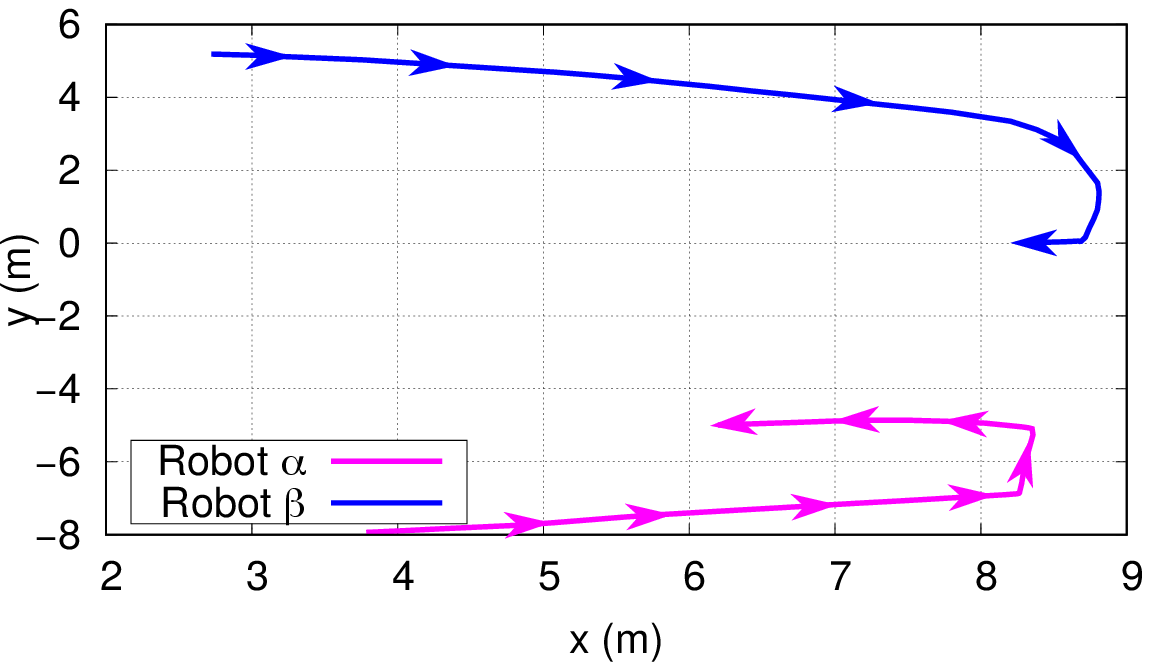}
        }
        \subfigure[Mean residual for Case \Romannum{1}]{
\label{fig:heating_map_case1}
        \includegraphics[width=0.23\textwidth]{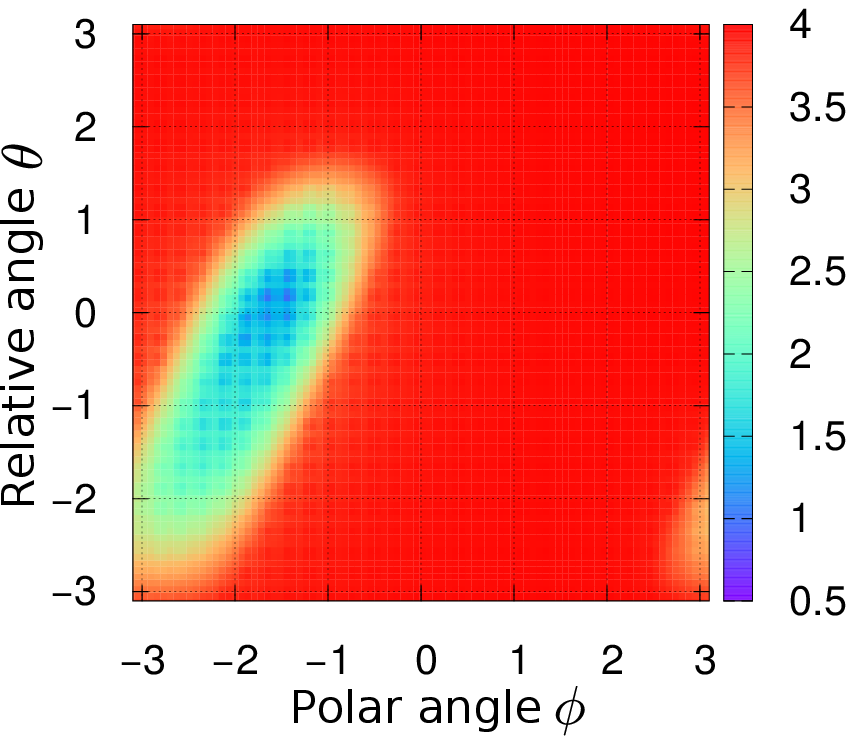}
        }
\hspace{-0.1in}\subfigure[Mean residual for Case \Romannum{2}]{
\label{fig:heating_map_case2}
        \includegraphics[width=0.23\textwidth]{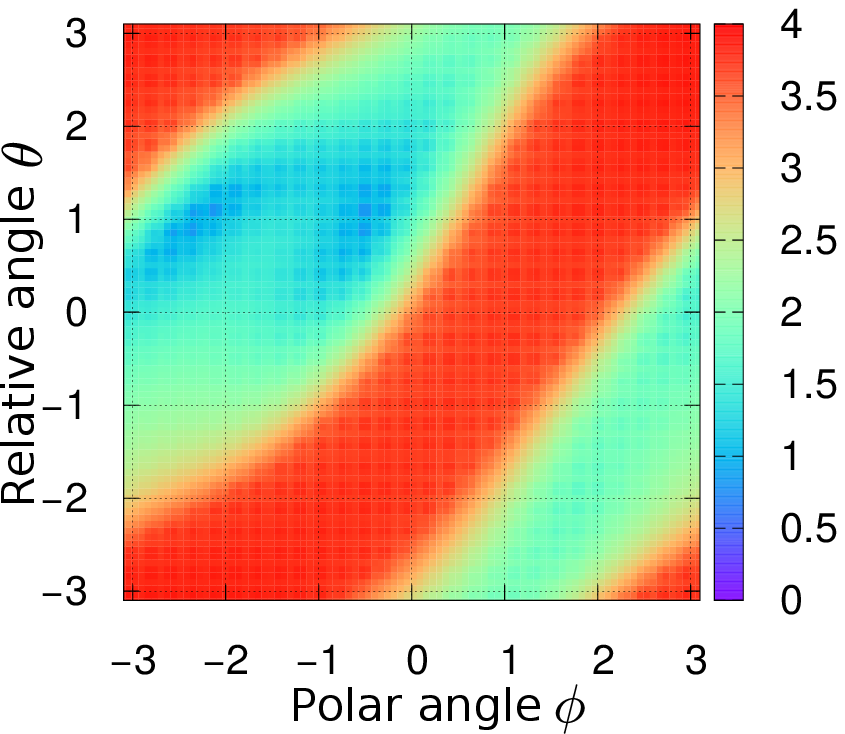}
        }
   \caption[]{Trajectories of two test cases with a window size $\tau=50$ and the mean residual error for different polar angle $\phi$ and relative angle $\theta$. The global minimal can be found by an arbitrary initial guess for Case \Romannum{1}, however this is not guaranteed in Case \Romannum{2}.}
\label{coarse_search_heatmap}
\vspace{-0.5cm}
\end{figure}

We propose to use a coarse search strategy to find a decent approximation of relative pose as initial guess for optimization in Equation \ref{eq:optimization}.
Based on coarse guess, we then perform nonlinear optimization with Equation \ref{eq:optimization}.
As UWB ranging is considered to be accurate, robot $\beta$ will approximately located on a circle centered at robot $\alpha$ with a radius $r_{\alpha,\beta}^t$ (known by UWB ranging) and an angle $\phi$ (to be estimated) in polar coordinate system. 
Therefore, instead of searching the whole 2D space with Equation \ref{eq:optimization}, 
we search in a polar space with angle between $-\pi$ and $\pi$ given the known UWB ranging.
The angular searching step size for $\theta$ and $\phi$ is denoted as $\delta$.
The number of searching steps (i.e., $w_\phi$ and $w_\theta$) for $\theta$ and $\phi$ are denoted as $w_\phi=w_\theta=\lceil \frac{\pi}{\delta} \rceil$.
We define the entire searching space as $\overline{w}=\{-w_\phi,w_\phi\} \times \{-w_\theta,w_\theta\}$.
The optimization of Equation \ref{eq:optimization} is approximated to find the minimum residual given ${ \widetilde{\x}_{\alpha, \beta}^t} \in \mathbf{w}$:
\begin{equation}
\small
 \begin{split}
 \mathbf{w}=\{ (\underbrace{r_{\alpha,\beta}^t\cos(\delta i_\phi)}_{\text{Position $x$} }, \underbrace{r_{\alpha,\beta}^t\sin(\delta i_\phi)}_{\text{Position $y$}}, \delta i_\theta): (i_\phi, i_\theta) \in \overline{w} \}
 \end{split}
\end{equation}
An overview of the algorithm is shown in Algorithm \ref{searching_algorithm}. 
The residual errors with respect to different $\phi$ and $\theta$ for two different test cases are shown in Figure \ref{coarse_search_heatmap}. 
The figure shows the residual follows a multimodal distribution, and a global minimum might be not found from an arbitrary initial guess (for example the Case \Romannum{2} in Figure \ref{coarse_search_heatmap}).
The coarse solution ${ \widetilde{\x}_{\alpha, \beta}^t}$ is used as initial guess to optimize Equation \ref{eq:optimization} via g2o \cite{g2o}. 
The node in the graph is denoted as the pose to be estimated (i.e., $\overline{\x}_{\alpha, \beta}^{t}$) and the constraints are represented by the UWB ranging measurements.
The algorithm turns out to find the best configuration of the pose to satisfy the UWB ranging constraints based on maximum likelihood estimation.

\subsection{Module 2: Distributed Outlier Rejection}
\label{rejection_pcm}
\begin{figure}
\centering
\includegraphics[width=0.4\textwidth]{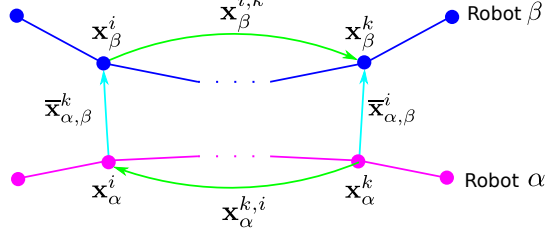}
\caption{Illustration of the distributed PCM consistency check performed by robot $\alpha$.
Robot $\alpha$ estimates the pose of robot $\beta$ at time $k$ and time $i$ using distributed pose estimation described in Section \ref{pose_est_ranging}. The estimated pose and odometry are passed for consistency check.}
\label{fig:pcm}
\vspace{-0.5cm}
\end{figure}
The pose estimation accuracy in Section \ref{pose_est_ranging} highly depends on UWB ranging accuracy and paths of robots.
These factors are not controllable for a practical applications, especially during autonomous navigation of robot in unknown environment with non-line-of-sight conditions.
Therefore, we use state-of-the-art outlier rejection algorithm, namely Pairwise Consistency Maximization (PCM)\,\cite{pcm}, to deal with suspicious pose estimations (i.e., loop closures) and tailor it to a distributed and online fashion.
Given two loop closures $\overline{\x}_{\alpha,\beta}^{k}$ and $\overline{\x}_{\alpha,\beta}^{i}$ between robot $\alpha$ and $\beta$,
the following equation is used to determine if two loop closures are pairwise consistent: 
\begin{equation}
\small
 \begin{split}
||\x_{\alpha}^{k,i} \oplus \overline{\x}_{\alpha, \beta}^{i} \oplus \x_{\beta}^{i,k} \ominus \overline{\x}_{\alpha, \beta}^{k} ||_{\Sigma} \le \chi^2_{\epsilon},
 \end{split}
 \label{eq:pcm}
\end{equation} 
where $||\cdot||_\Sigma$ represents the Mahalanobis distance. 
$\oplus$ and $\ominus$ denote composition and inversion operators of two poses \cite{compose_inversion}.
An illustration of consistency check is shown in Figure \ref{fig:pcm}. 
In ideal conditions, where all pose estimations are without noise, the composition of the poses along the cycle, which is formed by the two odometries (green lines in Figure \ref{fig:pcm}) and two loop closures (cyan lines in Figure \ref{fig:pcm}) is identify. 
Equation \ref{eq:pcm} evaluates if the error accumulated along the cycle is consistent with the noise covariance $\Sigma$.
$\chi^2_\epsilon$ is the quantile of the Chi-squared test for a given significance level $\epsilon$ \cite{doorslam}.

After consistency check by Equation \ref{eq:pcm}, the PCM builds an adjacency matrix to record the pairwise consistency of the loop closures.
The robot computes the maximum clique from the matrix, which gives the largest set of pairwise consistent loop closures (i.e., inliers).
To reduce computational cost for online operations,
we use an efficient version of PCM \cite{Kimera_Multi}, which incrementally updates the consistent loop closures with a heuristic search strategy to find the maximum clique. 
The inliers are passed to the pose graph for optimization, which will be detailed in Section \ref{dpgo}.

\subsection{Module 3: Distributed Pose Graph Optimization}
\label{dpgo}
Early multi-robot SLAM solutions focus on the centralized optimization.
The computation increases with the increase of robots, 
making this solution challenging in large environment with a large number of robots. 
In distributed SLAM \cite{gauss_seidel_SLAM}, each robot solves its pose graph optimization based on local information. 
The robots exchange their estimations to iteratively solve the global optimization problem.
We follow the approach proposed in \cite{dpgo}.
Each robot runs a DPGO module for trajectory optimization based on the odometry and the inlier loop closures. 
Particularly, DPGO performs local optimization by Riemann gradient descent. 
It solves a rank-restricted relaxation of pose graph optimization in a distributed manner.
During the update, the algorithm only requires the exchange of the poses connecting the loop closures and therefore 
saves a large amount of communication resources by avoiding the exchange of all pose information. 

Communication between robots is crucial in distributed SLAM system. 
The exchange of odometry is required in two modules, namely distributed pose estimation module in Section \ref{pose_est_ranging} and distributed outlier rejection in Section \ref{rejection_pcm}. 
The former shares short-term odometry measurements when two robots are in close proximity. 
The latter uses the relative pose estimation and its associated odometry for consistency check. Our solution is particularly suitable for the applications in ad hoc networks where external communication infrastructure is not deployed. 
For ad hoc networks, the robots are moving freely to form dynamic network topology, which leads to a limited communication range. 
Our implementation exchanges the updated poses and loop closures when the robots are in close proximity, 
while the communication is not required when the two robots are out of the UWB ranging.

\section{Experimental Results}
\label{sec:experimental_results}
\subsection{Experimental Setups}
We present experimental setups to demonstrate the proposed approach with three robots in an indoor environment with a size of 10m$\times$12m. 
In our setup, each robot carried one UWB node (Nooploop LinkTrack\footnote[1]{https://www.nooploop.com/}) with a maximum range up to 100 meters.
The sampling rate of UWB is set to 50Hz. 
The robot outputs the odometry measurements (differential drive with encoders) with a frequency of 10Hz. 
Three robots were manually controlled to move along different paths.
\begin{figure}
\centering
\includegraphics[width=0.4\textwidth]{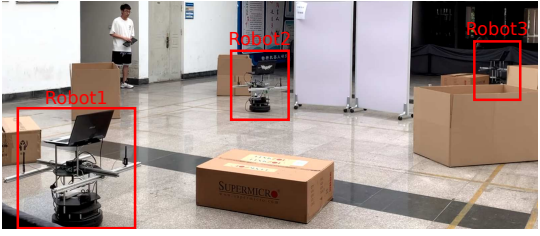}
\caption{Experiments with three TurtleBots moving in an indoor environment.}
\label{fig:robot}
\vspace{-0.3cm}
\end{figure}

\begin{table*}[]
\small
\caption{Relative pose estimation accuracy (translational error in meters and rotational error in degrees) and the time consumption (in millisecond) with various time window $\tau$ (in seconds) and different approaches.}
\centering

\begin{tabular}{|c|ccc|ccc|ccc|}
\hline
\multirow{2}{*}{$\tau$} & \multicolumn{3}{c|}{NLS optimization\cite{relative_pose_NLS}}                        & \multicolumn{3}{c|}{Coarse search}                                         & \multicolumn{3}{c|}{Proposed approach (coarse search+opt.)}                            \\ \cline{2-10} 
                     & \multicolumn{1}{c|}{Trans. error} & \multicolumn{1}{c|}{Rot. error} & Time & \multicolumn{1}{c|}{Trans. error} & \multicolumn{1}{c|}{Rot. error} & Time & \multicolumn{1}{c|}{Trans. error} & \multicolumn{1}{c|}{Rot. error} & Time \\ \hline \hline
$\tau=$10                   & \multicolumn{1}{c|}{8.19$\pm$6.85}      & \multicolumn{1}{c|}{70.71$\pm$53.85}    & 2ms & \multicolumn{1}{c|}{5.03$\pm$5.19}      & \multicolumn{1}{c|}{54.56$\pm$50.74}    & 4ms & \multicolumn{1}{c|}{4.54$\pm$5.24}      & \multicolumn{1}{c|}{49.34$\pm$50.61}    & 6ms \\ \hline
$\tau=$25                   & \multicolumn{1}{c|}{7.47$\pm$6.89}             & \multicolumn{1}{c|}{63.48$\pm$56.47}           &  4ms    & \multicolumn{1}{c|}{2.89$\pm$3.38}             & \multicolumn{1}{c|}{31.82$\pm$41.97}           &  5ms    & \multicolumn{1}{c|}{2.42$\pm$3.37}             & \multicolumn{1}{c|}{27.08$\pm$41.97}           &  9ms    \\ \hline
$\tau=$50                   & \multicolumn{1}{c|}{6.12$\pm$6.43}             & \multicolumn{1}{c|}{54.11$\pm$56.25}           &  6ms    & \multicolumn{1}{c|}{1.88$\pm$2.21}             & \multicolumn{1}{c|}{15.58$\pm$23.39}           &  7ms    & \multicolumn{1}{c|}{1.38$\pm$2.01}             & \multicolumn{1}{c|}{12.66$\pm$24.15}           &  13ms    \\ \hline
$\tau=$100                  & \multicolumn{1}{c|}{4.53$\pm$5.74}             & \multicolumn{1}{c|}{37.16$\pm$48.49}           &  10ms    & \multicolumn{1}{c|}{1.31$\pm$1.45}             & \multicolumn{1}{c|}{9.23$\pm$11.07}           &  9ms    & \multicolumn{1}{c|}{0.83$\pm$1.17}             & \multicolumn{1}{c|}{6.74$\pm$11.49}           &  19ms    \\ \hline
$\tau=$200                  & \multicolumn{1}{c|}{3.97$\pm$5.49}             & \multicolumn{1}{c|}{32.56$\pm$46.67}           &   15ms   & \multicolumn{1}{c|}{1.25$\pm$1.49}             & \multicolumn{1}{c|}{7.90$\pm$9.67}           &   14ms   & \multicolumn{1}{c|}{0.82$\pm$1.24}             & \multicolumn{1}{c|}{6.29$\pm$11.09}           &  29ms    \\ \hline
\end{tabular}
\label{table:different_window_size}
\end{table*}

\begin{figure}
  \centering
  \subfigure[Average translational error under various settings of $\delta$]{
\label{fig:trans_error_wrt_delta}
        \includegraphics[width=0.4\textwidth]{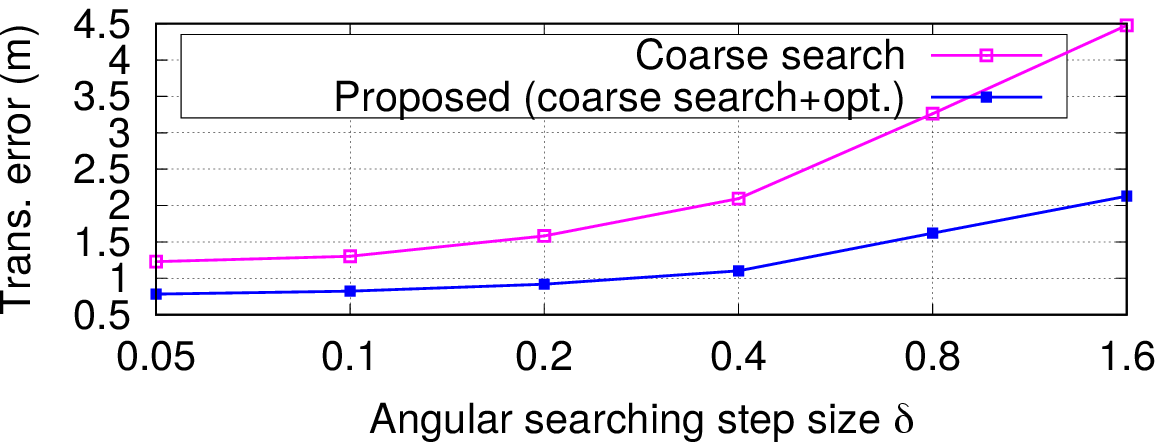}
        }
        \subfigure[Average rotational error under various settings of $\delta$]{
\label{fig:rot_error_wrt_delta}
        \includegraphics[width=0.4\textwidth]{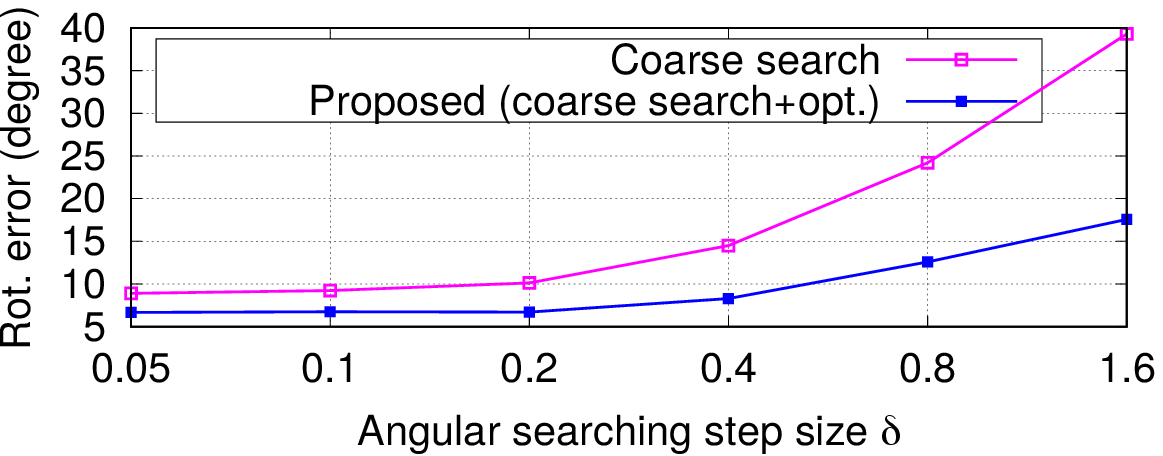}
        }
        \subfigure[Average time consumption under various settings of $\delta$]{
\label{fig:time_wrt_delta}
        \includegraphics[width=0.4\textwidth]{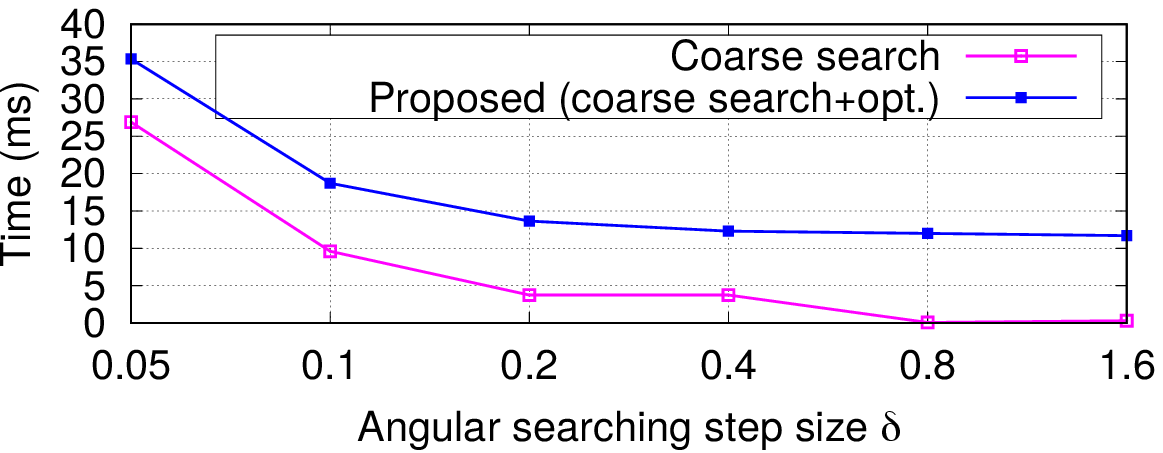}
        }
   \caption[]{Relative pose estimation accuracy and time consumption under different angular searching step sizes $\delta$.}
\label{figure:different_angular_searching_size}
\vspace{-0.3cm}
\end{figure}

All modules are running under ROS (Robot Operating System).
During the experiment, the maximum velocity of the robot was 0.2m/s.
The total distances travelled by the three robots were 63.5m, 60.6m, and 57.1m, respectively. To obtain the ground truth, Hokuyo LiDARs were installed to perform adaptive Monte Carlo localization\footnote[2]{http://wiki.ros.org/amcl} given a map created by GMapping. 
The number of poses collected by the three robots for evaluation were 314, 363, and 333, respectively.
An overview of the experimental snapshot is shown in Figure \ref{fig:robot}.
The relative translational and rotational errors are computed by comparing the estimated poses with the ground truth.
The mean squared error (MSE) in translation and rotation are computed as a metric to evaluate the positioning accuracy.

\begin{figure}
  \centering
  \subfigure[Average translational error under various ranging setting and ranging noise]{
\label{figure:max_range_tran}
        \includegraphics[width=0.4\textwidth]{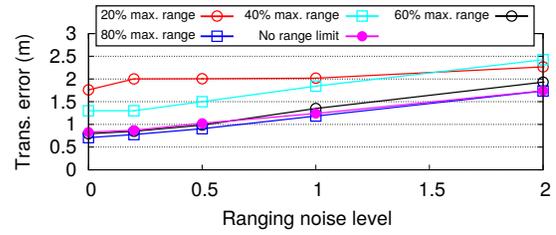}
        }
        \subfigure[Average rotational error under various ranging setting and ranging noise]{
\label{figure:max_range_rot}
        \includegraphics[width=0.4\textwidth]{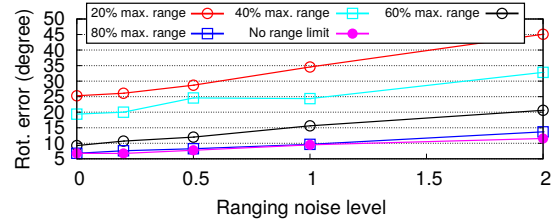}
        }
   \caption[]
{Relative pose estimation accuracy with respect to UWB maximum range and different UWB ranging noise.}
\label{figure:max_range}
\vspace{-0.2cm}
\end{figure}

\begin{figure}
  \centering
  \subfigure[Average translational error under various odometry noise]{
\label{figure:odom_level_tran}
        \includegraphics[width=0.4\textwidth]{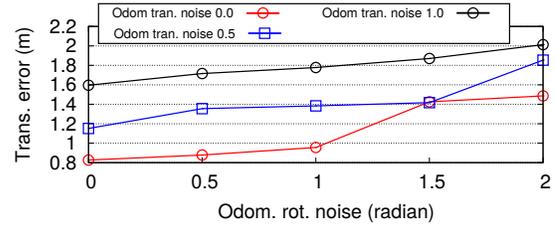}
        }
        \subfigure[Average rotational error under various odometry noise]{
\label{figure:odom_level_rot}
        \includegraphics[width=0.4\textwidth]{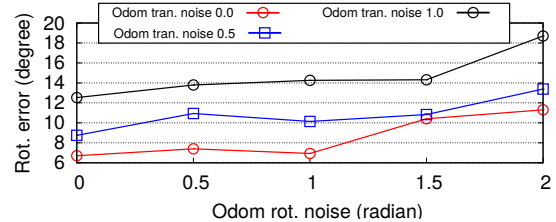}
        }
   \caption[]
{Relative pose estimation accuracy with respect to odometry noise.}
\label{figure:odom_level}
\vspace{-0.3cm}
\end{figure}

\begin{figure*}
  \centering
  \subfigure[Pose estimation of Robot2 and Robot3 with respect to Robot1 using short-term UWB and odometry measurements]{
\label{figure:track_raw}
        \includegraphics[width=0.3\textwidth]{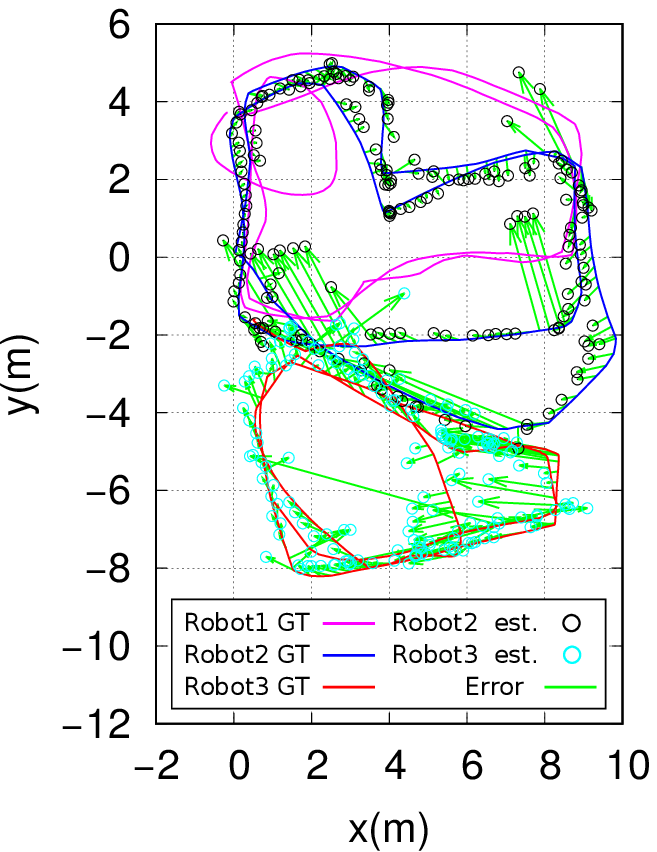}
        }\hspace{0.2in}\subfigure[Inliers by performing distributed PCM consistency check]{
\label{figure:track_pcm}
        \includegraphics[width=0.3\textwidth]{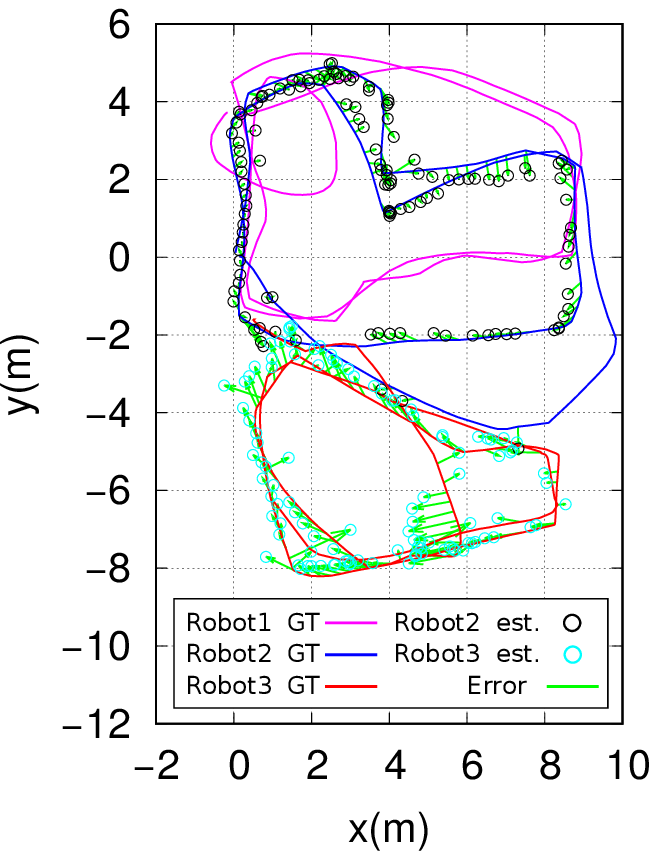}
        }\hspace{0.2in}\subfigure[Distributed pose graph optimization after PCM check]{
\label{figure:track_pgo}
        \includegraphics[width=0.3\textwidth]{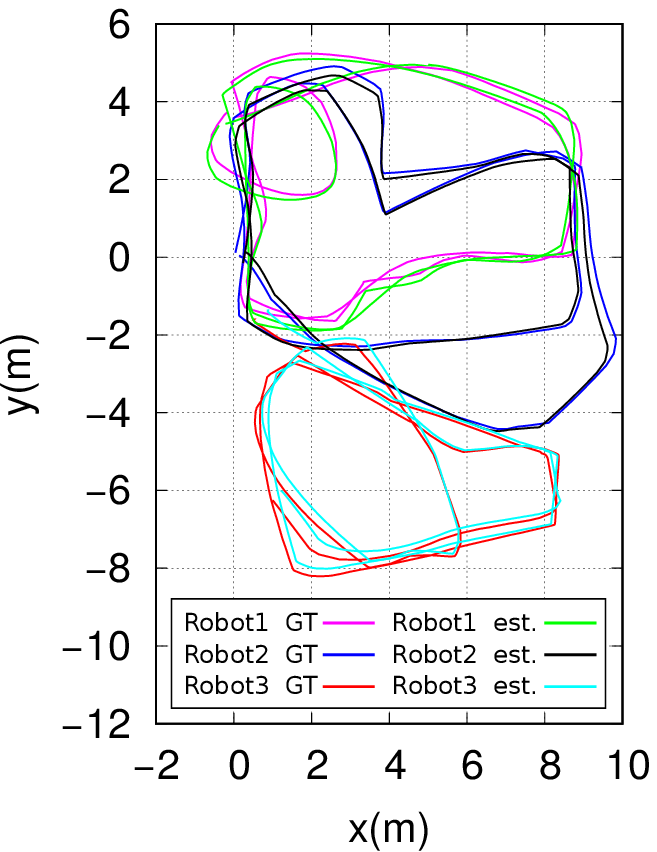}
        }
   \caption[]
{Trajectories estimated by different approaches. (a) Pose estimation using short-term UWB and odometry measurements; (b) Filtered poses after performing PCM check; (c) Optimized trajectory using distributed pose graph optimization.}
\label{figure:trajectory}
\vspace{-0.3cm}
\end{figure*}

\begin{table*}[]
\small
\caption{Pose estimation accuracy (translational error in meters and rotational error in degrees) with PCM check at different significance levels $\epsilon$ and evaluation of distributed pose graph optimization.}
\centering
\begin{tabular}{|c|cccc|cc|}
\hline
\multirow{2}{*}{$\tau$}     & \multicolumn{4}{c|}{Distributed PCM consistency check}                                                                              & \multicolumn{2}{c|}{Distributed pose graph optimization} \\ \cline{2-7} 
                            & \multicolumn{1}{c|}{$\epsilon$ in PCM} & \multicolumn{1}{c|}{No. of inliers} & \multicolumn{1}{c|}{Trans. error}  & Rot. error      & \multicolumn{1}{c|}{Trans. error}     & Rot. error       \\ \hline
\multirow{6}{*}{$\tau$=10}  & \multicolumn{1}{c|}{Without PCM}       & \multicolumn{1}{c|}{999}            & \multicolumn{1}{c|}{4.54$\pm$5.24} & 49.34$\pm$50.61 & \multicolumn{1}{c|}{2.13$\pm$1.12}    & 8.72$\pm$5.27    \\ \cline{2-7} 
                            & \multicolumn{1}{c|}{$\epsilon$=0.01}   & \multicolumn{1}{c|}{391}            & \multicolumn{1}{c|}{1.57$\pm$1.31} & 12.15$\pm$9.98  & \multicolumn{1}{c|}{0.78$\pm$0.45}    & 6.08$\pm$3.60    \\ \cline{2-7} 
                            & \multicolumn{1}{c|}{$\epsilon$=0.05}   & \multicolumn{1}{c|}{269}            & \multicolumn{1}{c|}{1.37$\pm$1.07} & 9.59$\pm$8.35   & \multicolumn{1}{c|}{0.79$\pm$0.43}    & 6.52$\pm$4.07    \\ \cline{2-7} 
                            & \multicolumn{1}{c|}{$\epsilon$=0.1}    & \multicolumn{1}{c|}{212}            & \multicolumn{1}{c|}{1.12$\pm$0.82} & 8.09$\pm$6.22   & \multicolumn{1}{c|}{0.61$\pm$0.36}    & 5.19$\pm$3.18    \\ \cline{2-7} 
                            & \multicolumn{1}{c|}{$\epsilon$=0.2}    & \multicolumn{1}{c|}{152}            & \multicolumn{1}{c|}{0.97$\pm$0.82} & 6.47$\pm$4.54   & \multicolumn{1}{c|}{0.71$\pm$0.43}    & 5.69$\pm$3.56    \\ \cline{2-7} 
                            & \multicolumn{1}{c|}{$\epsilon$=0.5}    & \multicolumn{1}{c|}{65}             & \multicolumn{1}{c|}{0.57$\pm$0.42} & 5.68$\pm$3.40   & \multicolumn{1}{c|}{0.53$\pm$0.36}    & 5.18$\pm$3.38    \\ \cline{2-7}
                            & \multicolumn{1}{c|}{$\epsilon$=0.8}    & \multicolumn{1}{c|}{25}             & \multicolumn{1}{c|}{0.73$\pm$0.72} & 7.97$\pm$6.02   & \multicolumn{1}{c|}{0.69$\pm$0.38}    & 6.60$\pm$3.96    \\ \hline
\multirow{6}{*}{$\tau$=50}  & \multicolumn{1}{c|}{Without PCM}       & \multicolumn{1}{c|}{999}            & \multicolumn{1}{c|}{1.38$\pm$2.01} & 12.66$\pm$24.15 & \multicolumn{1}{c|}{0.74$\pm$0.46}    & 5.42$\pm$4.05    \\ \cline{2-7} 
                            & \multicolumn{1}{c|}{$\epsilon$=0.01}   & \multicolumn{1}{c|}{904}            & \multicolumn{1}{c|}{0.95$\pm$0.91} & 6.93$\pm$6.75   & \multicolumn{1}{c|}{0.63$\pm$0.42}    & 4.78$\pm$3.50    \\ \cline{2-7} 
                            & \multicolumn{1}{c|}{$\epsilon$=0.05}   & \multicolumn{1}{c|}{863}            & \multicolumn{1}{c|}{0.90$\pm$0.83} & 6.14$\pm$5.16   & \multicolumn{1}{c|}{0.60$\pm$0.37}    & 4.29$\pm$3.07    \\ \cline{2-7} 
                            & \multicolumn{1}{c|}{$\epsilon$=0.1}    & \multicolumn{1}{c|}{782}            & \multicolumn{1}{c|}{0.76$\pm$0.59} & 5.78$\pm$4.30   & \multicolumn{1}{c|}{0.52$\pm$0.33}    & 4.43$\pm$3.00    \\ \cline{2-7} 
                            & \multicolumn{1}{c|}{$\epsilon$=0.2}    & \multicolumn{1}{c|}{663}            & \multicolumn{1}{c|}{0.64$\pm$0.42} & 5.25$\pm$3.71   & \multicolumn{1}{c|}{0.49$\pm$0.32}    & 4.18$\pm$2.83    \\ \cline{2-7} 
                            & \multicolumn{1}{c|}{$\epsilon$=0.5}    & \multicolumn{1}{c|}{394}            & \multicolumn{1}{c|}{0.51$\pm$0.31} & 4.52$\pm$2.99   & \multicolumn{1}{c|}{0.45$\pm$0.30}    & 4.01$\pm$2.81    \\ \cline{2-7}
                            & \multicolumn{1}{c|}{$\epsilon$=0.8}    & \multicolumn{1}{c|}{131}            & \multicolumn{1}{c|}{0.53$\pm$0.36} & 4.35$\pm$3.78   & \multicolumn{1}{c|}{0.52$\pm$0.26}    & 4.23$\pm$2.75    \\ \hline
\multirow{6}{*}{$\tau$=100} & \multicolumn{1}{c|}{Without PCM}       & \multicolumn{1}{c|}{999}            & \multicolumn{1}{c|}{0.83$\pm$1.17} & 6.74$\pm$11.49  & \multicolumn{1}{c|}{0.60$\pm$0.41}    & 5.05$\pm$3.56    \\ \cline{2-7} 
                            & \multicolumn{1}{c|}{$\epsilon$=0.01}   & \multicolumn{1}{c|}{987}            & \multicolumn{1}{c|}{0.75$\pm$0.76} & 5.88$\pm$4.57   & \multicolumn{1}{c|}{0.59$\pm$0.40}    & 4.96$\pm$3.44    \\ \cline{2-7} 
                            & \multicolumn{1}{c|}{$\epsilon$=0.05}   & \multicolumn{1}{c|}{974}            & \multicolumn{1}{c|}{0.74$\pm$0.74} & 5.71$\pm$4.15   & \multicolumn{1}{c|}{0.57$\pm$0.37}    & 4.88$\pm$3.34    \\ \cline{2-7} 
                            & \multicolumn{1}{c|}{$\epsilon$=0.1}    & \multicolumn{1}{c|}{929}            & \multicolumn{1}{c|}{0.69$\pm$0.63} & 5.42$\pm$3.71   & \multicolumn{1}{c|}{0.53$\pm$0.34}    & 4.71$\pm$3.22    \\ \cline{2-7} 
                            & \multicolumn{1}{c|}{$\epsilon$=0.2}    & \multicolumn{1}{c|}{736}            & \multicolumn{1}{c|}{0.64$\pm$0.49} & 5.19$\pm$3.48   & \multicolumn{1}{c|}{0.53$\pm$0.34}    & 4.69$\pm$3.29    \\ \cline{2-7} 
                            & \multicolumn{1}{c|}{$\epsilon$=0.5}    & \multicolumn{1}{c|}{702}            & \multicolumn{1}{c|}{0.51$\pm$0.34} & 5.47$\pm$3.42   & \multicolumn{1}{c|}{0.48$\pm$0.32}    & 4.92$\pm$3.32    \\ \cline{2-7}
                            & \multicolumn{1}{c|}{$\epsilon$=0.8}    & \multicolumn{1}{c|}{252}            & \multicolumn{1}{c|}{0.54$\pm$0.36} & 5.08$\pm$3.10   & \multicolumn{1}{c|}{0.49$\pm$0.31}    & 4.89$\pm$3.19    \\ \hline
\end{tabular}
\label{table:pcm_dpgo}
\vspace{-0.5cm}
\end{table*}

\subsection{Evaluation of Distributed Pose Estimation using Short-term UWB and Odometry}
\subsubsection{Impact of sliding window size $\tau$}
In the first set of experiments, we evaluated the performance of our pose estimation using UWB ranging and odometry measurements. 
We fixed the angular searching step size $\delta$ to 0.1 and examined the positioning accuracy and average time consumption for one pose estimation under the impact of sliding window size $\tau$, as shown in Table \ref{table:different_window_size}. 
This table also lists a comparison of the accuracy between nonlinear least square (NLS) optimization \cite{relative_pose_NLS}, coarse search-based approach, and the combination of coarse search and NLS optimization. 
As can be seen from Table \ref{table:different_window_size}, 
our approach, which combines the coarse search and NLS optimization, provides the best accuracy for all settings of $\tau$. 
For example, the combined approach gives a localization accuracy of 0.83m in translation and 6.74$\degree$ in rotation for $\tau$=100, which lead to improvements of 36.6\% and 27.0\% when compared with the pure coarse search-based approach (accuracy of 1.31m in translation and 9.23$\degree$ in rotation).
We also observed that the localization accuracy improves with the increase of the window size $\tau$, but at a cost of more computational resources.

\subsubsection{Impact under different angular searching steps $\delta$}
We evaluated the performance of relative pose estimation under various angular searching steps $\delta$.
We fixed the sliding window size $\tau$=100.
The positioning accuracy and the average computational time for one pose estimation are shown in Figure \ref{figure:different_angular_searching_size}.
As can be seen from this figure, for all parameter settings, optimization with a coarse initial guess improves the localization accuracy when compared to the pure coarse search-based approach.
The localization error is reduced with the decrease of the angular searching step, while the computational time increases at the same time.

\subsubsection{Impact of different UWB and odometry settings}
In this series of experiments, we set the UWB ranging limit (i.e., 20\%, 40\%, 60\%, and 80\% of the maximum UWB range reported during the test) to compare the accuracy.  
We also added Gaussian noise to raw UWB ranging to verify the performance under various noise levels.
We fixed the searching step size $\delta$=0.1 and the sliding window size $\tau$=100.
The result is shown in Figure \ref{figure:max_range}.
As can be seen from this figure, the localization accuracy is reduced with short UWB range. 
We also observed long UWB range in general helps to improve the localization accuracy due to the large number of UWB observations to support pose estimation. 
However, longer UWB ranging brings more uncertainty, particularly in non-line-of-sight conditions, which leads to a decrease of the localization accuracy. 
For all experimental settings, adding ranging noise will obviously lead to a decrease of localization accuracy.

We compared the localization accuracy with respect to various odometry errors. 
We added Gaussian noises to the position and heading of odometry and show the results in Figure \ref{figure:odom_level}. 
As can be seen from this figure, the accuracy decreases with the increase of odometry noise. 
However, we believe a good odometry can be provided by the on-board sensors of the robot, 
in particularly with the assistance of the state-of-the-art visual odometry or LiDAR odometry techniques.

\begin{table}[]
\small
\caption{Comparison between our approach and the literature (UWB SLAM in \cite{xielihua_uwb_collaborative}).}
\begin{tabular}{|c|c|c|c|}
\hline
Approach            & Pure odom. & Ours      & Literature\cite{xielihua_uwb_collaborative}  \\ \hline
Tran. error(m)     & 0.85$\pm$0.52  & 0.45$\pm$0.30 & 0.71$\pm$0.52 \\ \hline
Rot. error(degree) & 8.99$\pm$5.8   & 4.01$\pm$2.81 & 6.76$\pm$4.79 \\ \hline
\end{tabular}
\label{table:compare_to_range_SLAM}
\vspace{-0.7cm}
\end{table}

\subsection{Distributed Pose Graph Optimization with PCM Outlier Rejection}
In this series of experiments, we examined the localization accuracy with PCM check as well as the performance of the distributed pose graph optimization. 
We set the angular searching step size $\delta$=0.1.
The results are listed in Table \ref{table:pcm_dpgo}.
As can be seen from this table, PCM filters out the suspicious loop closures and improves the overall localization accuracy. 
For a sliding window size $\tau$=10, we obtain a localization accuracy of 1.37m in translation and 9.59$\degree$ in rotation with significance level $\epsilon=0.05$, which gives improvements of 69.8\% and 80.6\% when compared to the case without PCM check (4.54m in translation and 49.34$\degree$ in rotation).
In addition, we observe that the number of inliers decreases with the increase of the significance level $\epsilon$ in PCM.
A too large $\epsilon$ will harm the online operations, particularly during the early stage of trajectory estimation: 
potential good loop closures are pruned and the pose graph optimization has to be postponed until a reasonable number of loop closures are reached.
Meanwhile, a small $\epsilon$ is not able to distinguish the erroneous loop closures and results in an increase of the localization error.
The distributed pose graph optimization further improves the localization accuracy by the additional incorporation of odometry information.
For $\tau$=50 and $\epsilon$=0.5, we obtained a localization accuracy of 0.45m in translation and 4.01$\degree$ in rotation (the best accuracy achieved), with improvements of 11.8\% and 11.3\% when compared to the results without pose graph optimization (i.e., 0.51m in translation and 4.52$\degree$ in rotation).
The pose estimated with UWB and odometry, inliers after PCM check, and estimated trajectory using distributed pose graph optimization are visualized in Figure \ref{figure:trajectory}.

Table \ref{table:compare_to_range_SLAM} compared our approach with the conventional UWB SLAM \cite{xielihua_uwb_collaborative}, 
that uses UWB ranging for SLAM in a centralized manner.
As can be seen from this table, our approach outperforms the conventional UWB SLAM in localization accuracy. 
It is important to point out that the conventional UWB SLAM requires initial positions of the robots to be known publicly, while our approach discards this assumption by performing the relative pose estimation using local UWB ranging and odometry measurements. 

\subsection{Computation and Communication Efficiency}
Finally, we evaluated the computational cost and the communication resources involved in our approach, as listed in Table \ref{table:time_communication}. 
We choose the window size $\tau$=50 and the angular searching step size $\delta$=0.1.
The significance level in PCM is set to be $\epsilon$=0.1.
Our approach was tested on an AMD Ryzen 5950X processor with a frequency of 3.4GHz.
The computational time was cumulated over all three robots based on one pose update.
The pose is expressed in double precision (8 Bytes) in our implementation.
We show the total data communicated among all three robots at different updating frequencies (i.e., 10Hz and 1Hz) of DPGO.
As can be seen from this table, the total computational time required for the three modules is 81ms, which we believe is fast enough for the online localization and control of mobile robots. 
The cost of communication is small, which can be easily deployed in most robotics applications.
As can be also seen from this table, performing DPGO at a higher frequency costs more communication resources, due to the frequent exchange of the poses among the robots.

\begin{table}[]
\small
\caption{Computational time (millisecond, ms) and total communication cost (Megabytes, MB) for our approach.}
\centering
\begin{tabular}{|cccc|cc|}
\hline
\multicolumn{4}{|c|}{Time consumption}                                                                & \multicolumn{2}{c|}{Communication}   \\ \hline
\multicolumn{1}{|c|}{Pose est.} & \multicolumn{1}{c|}{Dist. PCM} & \multicolumn{1}{c|}{DPGO} & Total & \multicolumn{1}{c|}{10Hz}   & 1Hz    \\ \hline
\multicolumn{1}{|c|}{39ms}      & \multicolumn{1}{c|}{24ms}       & \multicolumn{1}{c|}{18ms}  & 81ms  & \multicolumn{1}{c|}{13.8MB} & 4.3MB \\ \hline
\end{tabular}
\label{table:time_communication}
\end{table}
\section{Conclusions}
\label{sec:conclusions}
We proposed an approach for distributed SLAM using UWB ranging and odometry measurements for a group of robots in unknown environments. 
The proposed method determines the relative pose between robots based on short-term UWB ranging and odometry measurements. 
In addition, we presented a distributed pose graph optimization with pairwise consistency check for trajectory estimation.
The proposed approach was verified with three robots in an indoor environment with a size of 10m$\times$12m.
The results showed that the proposed distributed ranging SLAM achieved a localization accuracy of 0.45m in translation and 4.01$^\circ$ in rotation.
Our approach provides a solution to the robotics community for the localization of a team of robots without any knowledge about the infrastructure. 
In future, we would like to generalize our approach to 3D, 
which requires the coarse search of the initial guess in 3D polar coordinate system.
The comparison to the quadratically constrained quadratic programming-based solution \cite{nguyen2022relative} will be also included in our future work.

\bibliographystyle{IEEEtran}

\bibliography{bibSpace}

\end{document}